\documentclass[12pt]{amsart}

\usepackage{common}
\usepackage{mathrsfs}

\newcommand{\dd}{\m{\bold d}}

\title{On sample complexity of neural networks}

\author{Alexander Usvyatsov}

\address{Alexander Usvyatsov\\
    Universidade de Lisboa \\
  Centro de Matem\'{a}tica e Aplica\cc{c}\~{o}es Fundamentais\\
  Av. Prof. Gama Pinto,2\\
  1649-003 Lisboa \\
  Portugal}

\thanks{
    The author thanks the Funda\cc{c}\~{a}o para a Ci\^{e}ncia e a Tecnologia for partial support
    of this research,
    grant no. SFRH / BPD / 34893 /
2007}

\date{\today}
\begin{document}

\begin{abstract}
  We consider functions defined by deep neural networks as definable objects in an o-miminal expansion of the real field, and derive an almost linear (in the number of weights) bound on sample complexity of such networks.
\end{abstract}

\maketitle

\section{Introduction and Preliminaries}

\subsection{Introduction}

  Recall that a function $f\colon \setR^k \to \setR$ is called \emph{restricted analytic} if there exists a function $\bar f\colon \setR^k\to\setR$ and a closed interval $[a.b]$ such that $\bar f$ is analytic in some neighborhood of $[a,b]^k$, and $f = \bar f$ on $[a,b]^k$ (and equals $0$ on the complement of $[a,b]^k$). Note that all activation functions of neural networks discussed in literature can be considered in this context.

  Let $\cC$ be a binary neural network with arbitrary restricted analytic activation functions. Note that we do not require that the activations functions at different nodes are all the same. Then $\cC$ defines a binary function $\cF = \cF(x_1,\ldots,x_n,w_1,\ldots,w_m: \setR^n\times \setR^m \to \set{0,1}$, where $x_1,\ldots,x_n$ are the inputs and $w_1,\ldots.w_m$ are the weights. Given a fixed collection of weights $\bar w = w_1,\ldots, w_m \in \setR$ we therefore obtain a binary function $\cF_{\w} \colon \setR^n \to \set{0,1}$.

  Consider the hypothesis class $\cH = \set{\cF_{\w}\colon \w \in \setR^m}$. It is well known  (e.g. \cite{Koiran96neuralnetworks,Bartlett_vapnik-chervonenkisdimension,Sontag98vcdimension,Karpinski95polynomialbounds}) that, depending on the activation functions of $\cC$, the VC-dimension of $\cH$ could be quadratic in $m$,even in quite simple and natural cases (e.g., linear activations, or a fixed sigmoid $\sigma$). This leads one to conclude that the best possible theoretical upper bound on sample complexity $k(\eps,\delta)$ of such $\cH$ is $O\left(\frac{m^2+\ln\frac 1\delta}{\eps}\right)$ for $(\eps,\delta)$-PAC learnability, or $O\left(\frac{m^2+\ln\frac 1\delta}{\eps^2}\right)$ for agnostic $(\eps,\delta)$-PAC learnability (see Theorem 6.8 in \cite{Shalev-shwartz_fromtheory}). In other words, for most activation functions used in practice, sample complexity of a neural network appears to be quadratic in the number of weights, and therefore $O(k^4)$ where $k$ is the size of $\cC$, i.e., the number of nodes in it. Moreover, for some non-algebraic, but still ``tame'', activations, such as $\sigma = \tanh$, VC-dimension is known to be $m^4$ (Karpinski and Macyntire \cite{Karpinski95polynomialbounds}); that is, the sample complexity  appears to be $O(k^8)$.

  However, it is intuitively clear that the number of ``degrees of freedom'' of $\cH$ is the number of weights, and not the square of the number of weights. One would therefore expect the sample complexity to be linear in $m$, hence $O(k^2)$ where $k$ is the size of $\cC$. VC-dimension does not, therefore, seem to explain this phenomenon. Even if we restrict ourselves to a very limited class of threshold activation functions, the VC-dimension of $\cH$ is still going to be $m\log(m)$. One way of settling the issue is simply noting, as in \cite{Livni:2014:CET:2968826.2968922}, that, since all real numbers involved in the computation of $\cF$ are in practice represented by a finite number of bits, one can without loss of generality restrict their attention to a subfamily of $\cH$ with a linear VC-dimension. This solution, however, still seems somewhat unsatisfying.

\smallskip

  In this note we observe that one can obtain a much better bound on sample complexity in terms of the number weights, once  the notion of VC-dimension is replaced with that of (combinatorial) VC-density. We shall recall that VC-density of any hypothesis set $\cH$ that arises from a neural network as above is $m$, and compute an upper bounds on sample complexity using combinatorial density. This will yield an $O(m\log(m)$ bound for any neural network $\fC$, provided that all the activation functions are restricted analytic. We will, however, have to pay a small price in dependence on either the confidence level $\delta$, or on the acceptable error $\eps$. This makes sense, since the bound $O\left(\frac{m^2+\ln\frac 1\delta}{\eps^2}\right)$ is known to be tight; however, the additional factor of $\log(1/\eps)$ seems insignificant in comparison with the gain ($m\log(m)$ as opposed to $m^2$ or even $m^4$), especially for very large networks used in practice today. We also hope that this factor can be improved further using a more careful analysis. In addition, we believe that using our approach and more sophisticated techniques, one can obtain a \emph{linear} dependence on $m$, which would fully settle the issue raised above. We will return to this in a future work. 

  Let us also note that more general activation function can be allowed in our analysis. As observed in \cite{Karpinski95polynomialbounds}, there are neural networks with a smooth activation function and infinite VC-dimension; in this case, by the general theory, VC-density will be infinite as well. However, one can allow \emph{certain} unrestricted functions: e.g., the exponential function $e^x$ (and, more generally, any function ``definable'' from $e^x$), or the function $x \mapsto x^{-1}$ which is defined to be $\frac{1}{x}$ for $x\neq 0$ and $0$ for $x=0$. In general, the only requirement that we have on the collection of all the activation functions of $\cC$ is that they are all \emph{simultaneously definable in a single o-minimal expansion of $\setR$}. In this context, this assumption is quite reasonable: all restricted analytic functions and $e^x$ are definable in $\setR_{exp,an}$, and $x \mapsto x^{-1}$ is definable in $(\setR_{an},{}^{-1})$; both of these structures are known to be o-minimal. In particular, the case of $\sigma=\tanh$ is also covered by our analysis. There are many references for o-minimality of various expansions of $\setR$, e.g., \cite{Wi,vDDMil,vDDSp,vDDSp2,vDDSp3}.

\smallskip

O-minimality has already had many fruitful applications in mathematics and computer science (for example, in verification and control theory, e.g. Brihaye \cite{Wallonie-bruxelles_verificationand}). Techniques from o-minimality have already been used in the study of neural networks, particularly, in computations of VC-dimension by Karpinsky and Macyntire \cite{Karpinski95polynomialbounds}. We believe that incorporating the progress of the last 20 years may lead to more illuminating results and yield new ideas and techniques. This note is just a small step in that direction.

\section{The setting} % (fold)
\label{sec:the_setting}

% section the_setting (end)
First we recall some basic notions from statistical learning theory. 

\subsection{VC-dimension} % (fold)
\label{sub:vc_dimension}

% subsection vc_dimension (end)

Let $\cX$ be a set. We denote by $2^{\cX}$ the power set of $\cX$. In our case, $\cX = \setR^n$.

Recall that the VC-dimension of the collection  of subsets $\cA\subseteq 2^{\cX}$ of $\cX$ is defined to be the maximal size (if exists) of a finite subset of $\cX$ which is shattered by $\cA$, i.e. 

$$\text{VC}(\cA) = \sup{|B|<\infty \colon B\subseteq \cX, |B\cap\cA|=2^{|B|}}$$

So if the maximum does not exist, we say that $\text{VC}(\cA)=\infty$.

In the definition above, $B\cap\cA = \set{B\cap A \colon A \in \cA}$. So $|B\cap\cA|=2^{|B|}$ if and only if for every subset $B' \subseteq B$ there exists $A' \in \cA$ so that $A'\cap B = B'$ (this is the origin of the term ``shattered''). Hence infinite VC-dimension means that $\cA$ shatters arbitrarily large sets (but not necessarily all sets). See e.g. Sontag \cite{Sontag98vcdimension} for more details and examples.

Given $\cA$ as above, $B\subseteq \cX$ finite and $B' \subseteq B$, we will say that $\cA$ \emph{recognizes} $B'$ in $B$ if for some $A' \in \cA$ we have $A'\cap B = B'$. So $\cA$ shatters $B$ if it recognizes all of its subsets.  

\smallskip

The relevance of VC-dimension to learning theory lies in the following simple but brilliant observation. It turns out that there is a sharp dichotomy in the number of subsets of an arbitrary set finite set $B$ that any collection $\cA$ can recognize. Specifically, either $\cA$ shatters arbitrary large sets (so $\text{VC}(\cA) = \infty$) or for any set large enough finite $B$, $\cA$ only recognizes a polynomial number of subsets of $B$. Moreover, if $\text{VC}(A) = d < \infty$, then for any finite $B \subseteq \cX$, the number of subsets of $B$ that $\cA$ can recognize is $O(n^d)$. This fact is known as the Sauer-Shelah Lemma, and it was proven independently by Sauer, Shelah, Perles, and Vapnik and Chervonenkis in slightly different contexts for different purposes around the same time. In other words, 

\begin{lem}\label{lem:sauer}(Sauer-Shelah Lemma) Let $\cA$ be a collection of subsets of a set $\cX$. Then either $\text{VC}(A) = \infty$, or, if $\text{VC}(A) = d < \infty$, then for every finite $B \subseteq \cX$ we have
$$ |B\cap \cA| = O(|B|^d) $$

\end{lem}

A more precise formula  can be given, but it is of no interest to us here. 

\subsection{VC-density} % (fold)
\label{sub:vc_density}

% subsection vc_density (end)
Motivated by the Sauer-Shelah Lemma, one can make the following definition:

\begin{dfn}\label{dfn:growth function}
  Let $\cX$, $\cA$ be as above. We define the \emph{growth function} of $\cA$, $\tau_{\cA}:\setN\to\setN$ as follows:
  $$ \tau_{\cA}(n) = \max{|B\cap \cA|\colon B\subseteq \cX, |B|=n}$$
\end{dfn}

In other words, $\tau_{\cA}(n)$ measures the maximal number of subsets of a set of size $n$ that $\cA$ can recognize. By the Sauer-Shelah Dichotomy Lemma, we have either $\tau_{\cA}(n) = 2^n$ or all $n$ (this case corresponds to infinite VC-dimension), or $\tau_{\cA}(n)$ is sub-polynomial, and in fact, $\tau_{\cA}(n) = O(n^d)$ where $d = \text{VC}(\cA)$. 

It is natural to ask whether the exponent $d$ above is optimal. And indeed, it turns out that in most cases it is really not. The ``true'' measure of the exponent in the growth function is called the \emph{combinatorial density} or the \emph{VC-density} of $\cA$, and it is denoted by $\text{vc}(A)$. More precisely:

\begin{dfn}\label{dfn:VCdensity}
  Let $\cX$, $\cA$ as above. Then the VC-density of $\cA$ is defined as follows:
  $$ \text{vc}(A) = \inf\set{q\in \setQ\colon \tau_{\cA}(n) = O(n^q)} $$

\end{dfn}

Note:

\begin{obs}\label{obs:basicproperties}
  \begin{enumerate}
    \item $\text{vc}(\cA)\le\text{VC}(\cA)$ for all $\cA$ [this somewhat explains the notation]
    \item $\text{vc}(\cA) = \infty$ if and only if $\text{VC}(\cA) = \infty$ for all $\cA$
  \end{enumerate}
\end{obs}

In general, VC-density is not particularly well-behaved. For instance, in Aschenbrenner et al \cite{Aschenbrenner11vapnik-chervonenkisdensity} examples of hypothesis classes of non-integer and even irrational VC-density are given. VC-density is also not known to be sub-additive (in \cite{KOU} Kaplan, Onshuus, and the author prove sub-additivity for a certain integer analogue of VC-density). However, in the particular examples that we are interested in, VC-density has been computed, and it turns out to be the minimal possible, as will be discussed in the next subsection.

\subsection{O-minimality} % (fold)
\label{sub:the_model_theoretic_framework}

% subsection the_model_theoretic_framework (end)

Let $(\cR,0,1,<,+,\cdot,f_\al \colon \al \in \cI)$ be an o-minimal expansion of $(\setR, 0,1,<,+,\cdot)$ with a collection of functions $\set{f_\al\colon \al \in \cI}$. For the purpose of this paper, $\cR$ can be simply $\setR_{an}$ or $\setR_{an,exp}$. In a nutshell, O-minimality means that any set, definable in $\cR$, is a finite collection of intervals. We refer to \cite{} for a survey on o-minimality an o-minimal expansions of $\setR$. 

\smallskip 

Let $\cC$ be a neural network with activation functions all definable in $\cR$. As described in the introduction, it defines a family of binary functions $\cH = \set{\cF_{\w}\colon \w \in \setR^m}$, which is precisely the hypothesis class that we are interested in, where each $\cF_{\w}$ is a boolean function on $\setR^n$. ALternatively, we can, of course, think of $\cF_{\w}$ as a subset $X_{\w}$of $\setR^n$ (say, the set of all $\x\in\setR^n$ on which $\cF_{\w}$ takes the value $1$). Moreover, the family $\set{X_{\w}\colon \w \in \setR^m}$ is \emph{uniformly definable} in $\cR$: there exists a formula, in fact, a \emph{quantifier free} formula, $\ph(\x,\w) = \ph(x,1,\ldots,x_n,w_1,\ldots,w_n)$ so that $X_{\w} = \set{\x \in \setR^n \colon \cR\models \ph(\x,\w)}$. 

In other words, our hypothesis class is exactly the class $\set{\ph(\x,\w)^{\cR^{n}} \colon \w \in \setR^m}$ where $\ph(\x,\w)^{\cR^{n}}$ is the set of ``solutions'' in $\cR^n$ of the formula $\ph(\x,\w)$ (where $\x$ are the variables, and $\w \in \setR^m$ is fixed). That is, $\ph(\x,\w)^{\cR^{n}} = \set{\x \in \setR^n \colon \cR\models \ph(\x,\w)}$.  See Karpinski and Macintyre \cite{Karpinski95polynomialbounds} for more details.

As in the previous subsection, we now denote this collection by $\cA$. So $\cA=\set{X_{\w}\colon \w \in \setR^m} = \set{\ph(\x,\w)^{\cR^{n}} \colon \w \in \setR^m}$

As mentioned in the introduction, depending on the activations of $\cC$ that we started with, it is possible that $\text{VC}(\cA) = m\log(m)$ or $m^2$ or $m^4$, or even $\infty$. Again, we refer to Sontag \cite{Sontag98vcdimension} for details. The finite possibilities can all be realized in the o-minimal setting that we are working in (examples in \cite{Sontag98vcdimension} can all be defined in $\setR_{an,exp}$), however, any uniformly definable family of sets in an o-minimal structure has a finite VC-dimension, therefore the last possibility is impossible in our case. In fact, this is true in a much wider class of dependent (NIP) structures (which we will not discuss here). 

However, a much stronger statement can be made concerning VC-density in an o-minimal structure. Specifically: 

\begin{thm}\label{thm:vcdensity in ominimal}
  Let $\cA$ be a uniformly definable family of sets in an o-minimal structure $\cR$. That is, assume that $\cA = \set{\ph(\x,\w)^{\cR^{n}} \colon \w \in \cR^m}$ for some formula $\ph(\x,\y) = \ph(x_1,\ldots,x_n,y_1,\ldots,y_m)$. Then $\text{vc}(\cA)\le m$.
\end{thm}

Note that one can not expect better: the simple formula $x=y_1 \lor x=y_2 \lor \ldots \lor x=y_m$ with one variable $x$ defines the family $\cA = \set{A \subseteq \cX \colon |A| \le m}$, so $|\cA\cap B|$ is all the subsets of $B$ of size at most $m$, hence roughly of the size $|B|^m$, at least for $B$ large enough. 

This theorem is due to Johnson and Laskowski \cite{Johnson09c.:compression}. It was obtained earlier for o-minimal expansions of the reals (which is the context we are considering) by Karpinski and Macyntyre \cite{Karpinski:1997:AVI:895414}. A more recent and general approach that applies in a much wider context can be found in \cite{Aschenbrenner11vapnik-chervonenkisdensity}. 

\section{Sample complexity I}

We now turn to computing the desired bound on sample complexity. In this section, we show an elementary computation, which provides a loose bound. 

We refer to Ben David and Shalev-Shwartz \cite{Shalev-shwartz_fromtheory} for basic concepts of statistical learning.

Let $\cH$ be a binary hypothesis class on a sample space $\cX$, and let $\cD$ be a probability distribution on $\cX$. Denote by $L_{\cD} \colon \cH\to[0,1]$ the 0-1 loss function function with respect to $\cD$. Essentially, $L_{\cD}(h)$ measures the propability 
(with respect to $\cD$) of a sample to be misclassified by $h$. See \cite{Shalev-shwartz_fromtheory} for details.  Recall that given a sample $S \in \cX^k$ of size $k$ and $h \in \cH$, we denote by $L_S(h)$ the 0-1 loss of $h$ with respect $S$. This is simply the percentage of elements of $S$ that $h$ misclassifies - the most natural estimate for $L_{\cD}(h)$, if all one sees is $S$). Recall also that $\cD^k$ denotes the product measure on $\cX^k$ that arises from $\cD$.

Recall that if we succeed in showing that, given $\eps,\delta>0$, there exists $k = k(\eps,\delta)$ such that with probability $1-\delta$ over the choice of a sample $S$ of size $k$, we have $\left|L_{\cD}(h)-L_S(h)\right| \le \eps$, then in particular this $k$ provides an upper bound on sample complexity for agnostic PAC learnability. In fact, such $k$ witnesses a stronger property called ``uniform convergence'' for $\cH$. 

The following basic fact is Theorem 6.11 in \cite{Shalev-shwartz_fromtheory}:

\begin{fct}\label{fct:theorem growth function}
   For every $h\in\cH$ and $\delta \in (0,1)$, with probability of at least $1-\delta$ over the choice of $S \sim \cD^m$ we have 
  \[
    \left|L_{\cD}(h)-L_S(h)\right| \le \frac{4+\sqrt{\log(\tau_{\cH}(2k)}}{\delta\sqrt{2k}}
  \]
 \end{fct} 

Where $\tau_{\cH}$ denotes the growth function of $\cH$, as defined in subsection \ref{sub:vc_density}.

\smallskip 

Now let us combine Fact \ref{fct:theorem growth function} with Theorem \ref{thm:vcdensity in ominimal}, i.e. recall that 
$\tau_{\cH}(2k) \le (2k)^m$:

\[
    \left|L_{\cD}(h)-L_S(h)\right| \le \frac{4+\sqrt{\log(2^mk^m)}}{\delta\sqrt{2k}} = \frac{4+\sqrt{m+m\log(k)}}{\delta\sqrt{2k}}
\]

% Note that if we manage to uniformly bound the expression on the left side of the last equation as a function of $k$, this would exactly prove the uniform convergence property for $\cH$, in particular yielding an upper bound on sample complexity of $\cH$.

Let $\eps>0$. We want the expression on the left side of the equation to be at most $\eps>0$. For this (assuming $k$ is large enough) it is enough to find $k$ such that

\[
    \frac{2\sqrt{\log(2^mk^m)}}{\delta\sqrt{2k}} = \frac{2\sqrt{m+m\log(k)}}{\delta\sqrt{2k}} \le \eps
\]

That is,

\[
\frac{4{m\log(2k)}}{\delta^2{2k}}\le \eps^2
\]

\[
\frac{\log(2k)}{{2k}}\le \frac{\eps^2\delta^2}{4m}
\]

Or

\[
\frac{{2k}}{\log(2k)}\ge \frac{4m}{\eps^2\delta^2}
\]

So we want $k$ large enough so that the following inequality holds:

\[
{2k}\ge \frac{4m}{\eps^2\delta^2}\log(2k)
\]

For this (see Lemma A.1 in \cite{Shalev-shwartz_fromtheory}) it is enough to have 

\[
  2k \ge 2\frac{4m}{\eps^2\delta^2}\log(\frac{4m}{\eps^2\delta^2})
\]

Or 

\[
  k \ge \frac{4m}{\eps^2\delta^2}\log(\frac{4m}{\eps^2\delta^2})
\]

We have therefore shown 

\begin{thm}\label{thm:sample complexity}
  A neural network with $m$ weights and activations simultaneously definable in some o-minimal structure admits the property of uniform convergence, and is therefore agnostic PAC-learnable, with sample complexity 

\[
  k = k(\eps,\delta) \le \frac{4m}{\eps^2\delta^2}\log(\frac{4m}{\eps^2\delta^2})
\]
\end{thm}

In the case of high VC-dimension, we get a much better dependence on $m$ at the expense of a worse dependence on $\delta$ than in the classical bounds. It is clear, however, that more careful analysis will yield better bounds. We confirm this in the next section. 

\section{Sample complexity II}

In this section we provide a tighter bound, significantly improving the dependence on $\delta$, at the expense of a worse multiplicative constant, and an additional factor of $\log(1/\eps)$. 

As in the previous section, we refer the \cite{Shalev-shwartz_fromtheory} for background, specifically, for the discussion of Rademacher complexity and its properties. 

The setting is the same as in the previous section: $\cH$ is a binary hypothesis class of VC-density $m$ on a sample space $\cX$, $\cD$ is a probability distribution on $\cX$; $L_{\cD} \colon \cH$ denotes the 0-1 loss function function with respect to $\cD$, $L_S(h)$ denotes the 0-1 loss of a hypothesis $h$ with respect to a sample $S \in \cX^k$.

Our main result is the following:

\begin{thm}\label{thm:sample complexity bound}
  Let $\cC$ be a binary neural network with $m$ weights and activation functions definable in an o-minimal structure. Let $\cH$ be the hypothesis set given by $\cC$. Then there exists a multiplicative constant $\hat C$ such that for every $\eps,\delta>0$ we have $\left|L_{\cD}(h)-L_S(h)\right| \le \eps$ with probability at least $1-\delta$ over the choice of $S$ for all $h \in \cH$, provided that 
  \[
    |S|\ge \hat C \left[\frac{m}{\eps^2}\log\left(\frac{2m}{\eps^2}\right) + \frac{\log\left(\frac{4}{\delta}\right)}{\eps^2}\right]
  \]
\end{thm}

\begin{prf}
Given a classification training set $\set{(x_i,y_i)\colon i\le k}$, let $\setA$ denote the set of all binary vectors in $\setR^k$ of the form $([h(x_1) = y_1],[h(x_2) = y_2],\ldots,[h(x_k) = y_k])$ for $h \in \cH$; where we naturally interpret the  value of a boolean expression as 1 if it is true, and 0 if it is false. By Fact \ref{fct:linear}, the size of $\setA$ is bounded by $Ck^m$ for some multiplicative constant $C$. Hence by Massart's Lemma (see Lemma 26.8 in \cite{Shalev-shwartz_fromtheory}), the Rademacher complexity of $\setA$ is bounded above by
\[
   \sqrt{\frac{2\log(Ck^m)}{k}} = \sqrt{\frac{2m\log(C'k)}{k}}
 \] 

For some constant $C'$ (and note that the norm of a binary vector in $\setR^k$ is at most $\sqrt{k}$, hence we get $\sqrt{k}$ in the denominator).

Now let $\delta>0$. By Theorem 26.5 in \cite{Shalev-shwartz_fromtheory}, with probability at most $1-\delta$, we get 

\[
  \left|L_{\cD}(h)-L_S(h)\right| \le \sqrt{\frac{8m\log(C'k)}{k}} + \sqrt{\frac{2\log(\frac{4}{\delta})}{m}} \le 2\sqrt{\frac{8m\log(C'k)+2\log(\frac{4}{\delta})}{m}}
\]

And so, given $\eps>0,\delta>0$, we need the sample size $k = k(\eps,\delta)$ to satisfy:

% \[
%   2\sqrt{\frac{8m\log(C'k)+2\log(\frac{4}{\delta})}{k}} \le \eps
% \]

% Or 

\[
  4{\frac{8m\log(C'k)+2\log(\frac{4}{\delta})}{k}} \le \eps^2
\]

Or

\[
  {\frac{k}{16m\log(C'k)+8\log(\frac{4}{\delta})}} \ge \frac{1}{\eps^2}
\]

Rewriting the desired inequality in the following form

\[
  {\frac{C'k}{8C'\left[2m\log(C'k)+\log(\frac{4}{\delta})\right]}} \ge  \frac{1}{\eps^2}
\]

We need 

\[
  {C'k} \ge  \frac{8C'\left[2m\log(C'k)+\log(\frac{4}{\delta})\right]}{\eps^2} = 16C'\frac{m}{\eps^2}\log(C'k) + 8C'\frac{\log(\frac{4}{\delta})}{\eps^2}
\]

Applying Lemma A.2 from \cite{Shalev-shwartz_fromtheory}, we can now find a constant $\hat C$ (which can be explicitly computed from $C'$, hence from $C$) so that the above inequality holds provided that 

\[
  k \ge \hat C \left[\frac{m}{\eps^2}\log\left(\frac{2m}{\eps^2}\right) + \frac{\log\left(\frac{4}{\delta}\right)}{\eps^2}\right]
\]

\end{prf}

This gives us a much better dependence on $\delta$ at the expense of the much less significant factor $\log(1/\eps)$ (and a worse multiplicative constant). It is possible that this factor can be improved using more sophisticated techniques; we intend to examine this question in future works. 

% \bigskip

% \[
%   \frac{4(m+m\log(k))}{\delta^2{2k}} \le \eps^2
% \]

% Or

% \[
%   \frac{4m(1+\log(k))}{\delta^2{2k}} \le \eps^2
% \]
% \[
%   \frac{(1+\log(k))}{{k}} \le \frac{2\eps^2\delta^2}{4m}
% \]
% \[
%   \frac{{k}}{(1+\log(k))} \ge \frac{4m}{2\eps^2\delta^2}
% \]
% \[
%   {{k}} \ge \frac{4m}{2\eps^2\delta^2}{(1+\log(k))}
% \]
% \[
%   {{k}} \ge \frac{4m}{2\eps^2\delta^2}{\log(k))}+\frac{4m}{2\eps^2\delta^2}
% \]
% \[
%   {{k}} \ge 4\frac{4m}{2\eps^2\delta^2}{\log(2\frac{4m}{2\eps^2\delta^2}))}+2\frac{4m}{2\eps^2\delta^2}
% \]    
% \[
%   {{k}} \ge 2\frac{4m}{\eps^2\delta^2}{\log(\frac{4m}{\eps^2\delta^2}))}+\frac{4m}{\eps^2\delta^2}
% \]  

\bibliography{common.bib}
\bibliographystyle{alpha}

\end{document}